\documentclass{article}

\usepackage{PRIMEarxiv}

\usepackage[numbers]{natbib}

\usepackage[utf8]{inputenc} % allow utf-8 input
\usepackage[T1]{fontenc}    % use 8-bit T1 fonts
\usepackage{hyperref}       % hyperlinks
\usepackage{url}            % simple URL typesetting
\usepackage{booktabs}       % professional-quality tables
\usepackage{amsfonts}       % blackboard math symbols
\usepackage{nicefrac}       % compact symbols for 1/2, etc.
\usepackage{microtype}      % microtypography
\usepackage{lipsum}
\usepackage{fancyhdr}       % header
\usepackage{graphicx}       % graphics
\graphicspath{{media/}}     % organize your images and other figures under media/ folder

\usepackage{amsmath,amssymb,amsfonts}

\usepackage{lineno}

\usepackage{graphicx}
\usepackage{textcomp}

\usepackage{xcolor}

\usepackage{paralist}
\usepackage{amsmath,amsfonts,amssymb,amsthm,dsfont}

\usepackage{subcaption}
\usepackage{hyperref}
\usepackage{url}
\usepackage{float}

\usepackage{multirow}
\usepackage{multicol}
\usepackage{booktabs}

\usepackage{algorithm}
\usepackage[noend]{algpseudocode}

\def\our{HyConEx}

\def\R{\mathbb{R}}

\def\lcal{\mathcal{L}}

\usepackage{hyperref}
\usepackage{color}

\title{\our{}: Hypernetwork classifier with counterfactual explanations for tabular data
}

\author{
    Patryk Marsza\l{}ek$^{1}$ \And
    Kamil Ksi\k{a}\.z{}ek$^{1}$ \And
    Oleksii Furman$^{2}$ 
    \And
    Ulvi Movsum-zada$^{1}$ \And
    Przemys\l{}aw Spurek$^{1, 3}$ \And
    Marek \'Smieja$^{1}$\thanks{marek.smieja@uj.edu.pl} \\ \\ 
    \textsuperscript{1}Faculty of Mathematics and Computer Science, Jagiellonian University, Cracow, Poland \\
    \textsuperscript{2}Wroc\l{}aw University of Science and Technology, Department of Artificial Intelligence, Wroc\l{}aw, Poland\\
    \textsuperscript{3}IDEAS Research Institute, Warsaw, Poland
}    
\begin{document}
\maketitle

\begin{abstract}
In recent years, there has been a growing interest in explainable AI methods. In addition to making accurate predictions, we also want to understand what the model's decision is based on. One of the fundamental levels of interpretability is to provide counterfactual examples explaining the rationale behind the decision and identifying which features, and to what extent, must be modified to alter the model's outcome. To address these requirements, we introduce \our{}, a classification model based on deep hypernetworks specifically designed for tabular data. Owing to its unique architecture, \our{} not only provides class predictions but also delivers local interpretations for individual data samples in the form of counterfactual examples that steer a given sample toward an alternative class. While many explainable methods generate counterfactuals for external models, there have been no interpretable classifiers simultaneously producing counterfactual samples so far. \our{} achieves competitive performance on several metrics assessing classification accuracy and fulfilling the criteria of a proper counterfactual attack. This makes \our{} a distinctive deep learning model, which combines predictions and explainers as an all-in-one neural network. The code is available at \url{https://github.com/gmum/HyConEx}. \textcolor{red}{This work has been published in \emph{Neurocomputing} and is available at: \href{https://doi.org/10.1016/j.neucom.2026.132748}{https://doi.org/10.1016/j.neucom.2026.132748}.}
\end{abstract}

% keywords can be removed

\section{Introduction}

Deep learning has made significant progress in the area of tabular data. It has been shown that recent deep learning models like TabPFN~\citep{hollmann2023tabpfn}, or VisTabNet~\citep{wydmanski2024vistabnet}, match or even exceed the performance of state-of-the-art gradient boosting ensembles, such as CatBoost~\citep{Dorogush2018CatBoostGB} or XGBoost~\citep{chen2016xgboost}. However, at the cost of increased performance, decisions made by deep learning models cannot be easily explained, which raises concerns regarding trust, accountability, and fairness, especially in critical applications. Instead of designing interpretable deep learning classifiers, which deliver explanations at the prediction stage, most explainable models operate post-hoc, i.e. they are applied externally to classification models to explain their decisions. Although such an approach is generally acceptable since it does not affect classification performance, it is not optimal because we need to construct multiple models, one operating on the output of the other. Moreover, every single explanation method requires time-consuming optimization.   

Explanations can be delivered at several levels. 
%However, we need not only high-performing classifiers but also models which can explain their decisions. 
For instance, banks, due to legal reasons, need to explain why they refuse or grant a loan. 
%Similarly, machine learning system supporting medical diagnostics should be able to explain why a particular diagnosis has been made and which treatment suits the most to the current patient state~\citep{Amann2020medicine,Barry2012decisions}. 
These explanations define the importance of input features, which are crucial for making a given decision. On the other hand, 
%Similarly, 
a bank customer would like to know what she/he should do to change the bank's decision~\citep{Guidotti2024counterfactual}. This topic is related to counterfactual explanations informing which features, and to what extent, have to be modified to obtain a desired model output, e.g. getting a loan. Feature importance methods deliver static information about classification (factual decision). In contrast, counterfactual examples are a type of causal reasoning and provide better insight into the dynamic of the classification rule.

While there exist highly accurate  machine learning classifiers for tabular data~\citep{Dorogush2018CatBoostGB,gorishniy2021revisiting,hollmann2023tabpfn,kadra2024interpretable}, as well as external methods for generating counterfactual explanations
%explaining the importance of particular features~\citep{Arik2021tabnet,Ribeiro2016lime,Lundberg2017shap}, or producing counterfactual explanations
~\citep{mothilal2020dice,Ribeiro2018anchors,Wielopolski2024ProbabilisticallyPC}, no existing model combines these two objectives. %The seminal work towards combining highly accurate classifier with interpretability aspect has been recently realized by Interpretable Mesomorphic Network (IMN) \citep{kadra2024interpretable}. In this deep learning approach a single hypernetwork returns a local linear model for every input example. While the linearity of the output model provides the importance score of features, the complexity of the model is hidden in the hypernetwork architecture.
In this paper, we make a step towards interpretable deep learning and combine the classification power of neural networks with the dynamic interpretability of counterfactual examples. We introduce \our{}, a type of hypernetwork, which for every input returns a local classifier equipped with \emph{counterfactual vectors} targeting all other classes. In this way, \our{} not only returns a highly accurate decision but also suggests possible options of how to change that decision. Unlike existing interpretable classifiers that provide feature importance scores explaining \emph{why} a decision was made, \our{} delivers counterfactual examples that answer \emph{how} to change that decision through actionable guidance generated in a single forward pass. Enclosing classification and counterfactual generation in one model allows us to speed-up the inference and avoid time-consuming optimization of counterfactuals generation, which is common for external counterfactual explainers \citep{Wielopolski2024ProbabilisticallyPC}. To the best of our knowledge, \our{} is the first deep learning approach having these two features enclosed in one model.

% \marek{Dopisac na czym polega nasza metody (konstrukcja)}

Our experiments demonstrate that \our{} achieves accuracy competitive to the state-of-the-art methods (Table~\ref{table:test_performances}). Moreover, it produces counterfactual explanations for all classes of comparable quality to external counterfactual explainers (Tables \ref{table:countefacts_2_class}, \ref{table:countefacts_multi_class} and \ref{table:countefacts_categorical}). Since counterfactuals are generated jointly with classification scores in a single forward pass, it does not introduce any additional computational overhead, which makes it a preferred choice over external methods that require time-consuming optimization.

\section{Related Works}
%%%%%%%%%%%%%%%%%%%%%%%%%%%%%%%%%%%%

In this section, we review the literature on interpretable classification, categorizing prior work into three main groups: classical interpretable models, post-hoc explanation techniques (including counterfactual methods) and hybrid approaches that integrate interpretability into the model design.

\subsection{Classical Interpretable Models}

Tabular data are ubiquitous in domains such as finance, medicine, and fraud detection. Traditional models such as logistic regression and decision trees have long been favored for their inherent transparency. Logistic regression \citep{hosmer2013lr}, for instance, has been extensively studied due to its linear formulation that directly relates feature coefficients to prediction outcomes. Similarly, linear Support Vector Machines (SVMs) \citep{cortes1995svm} offer interpretability through the explicit characterization of support vectors and decision boundaries. Decision trees provide a natural form of explanation by decomposing decisions into a sequence of simple, human-readable rules. However, while tree-based ensemble methods (for example, boosted trees such as XGBoost \citep{chen2016xgboost} and CatBoost \citep{Dorogush2018CatBoostGB}) achieve high predictive performance in tabular data, the aggregated nature of these models often obscures the interpretability of individual decisions, necessitating the use of additional explanation techniques.

\subsection{Post-Hoc and Counterfactual Explanations}

To mitigate the opacity of high-performing black-box models, several post-hoc explanation techniques have been proposed \citep{lakkaraju2020fool, ras2022explainable, zanjani2024explainable}. Methods such as Local Interpretable Model-agnostic Explanations (LIME)~\citep{Ribeiro2016lime} and SHapley Additive exPlanations (SHAP)~\citep{Lundberg2017shap} approximate complex models locally with simpler, interpretable surrogates that assign importance scores to individual features. These approaches have proven effective in providing insights into the local behavior of classifiers, albeit sometimes at the cost of stability and global consistency. 

Counterfactual explanation methods have emerged as a complementary paradigm by identifying minimal changes in input features that would alter a model prediction, thus offering actionable insights into how a different outcome might be achieved. These methods can be broadly categorized into endogenous explainers, which generate counterfactuals by selecting or recombining feature values from the existing dataset, and exogenous explainers, which produce counterfactual examples through interpolation or random data generation without the strict constraint of naturally occurring feature values \citep{Guidotti2024counterfactual}. Early work by Wachter et al. \citep{wachter2017counterfactual} laid the foundation for exogenous methods, and subsequent approaches have enhanced plausibility by imposing constraints such as convex density restrictions proposed by Artelt and Hammer \citep{artelt2020convex}, integrating density estimation via normalizing flows as in Probabilistically Plausibile Counterfactual Explanations using Normalizing Flows (PPCEF)~\citep{Wielopolski2024ProbabilisticallyPC}, or guiding the search for interpretable counterfactuals with class prototypes as in Counterfactual Explanations Guided by Prototypes (CEGP) \citep{van2021cegp}. On the endogenous side, the Contrastive Explanations Method (CEM) \citep{dhurandhar2018cem} emphasizes the importance of both pertinent positives and negatives to justify a prediction, an approach further refined by Feasible and Actionable Counterfactual Explanations (FACE)~\citep{poyiadzi2020face}, which identifies feasible and actionable modification paths that respect inherent data and decision constraints, and by the Case-Based Counterfactual Explainer (CBCE) \citep{keane2020cbce} that forms explanation cases by pairing similar instances with contrasting outcomes. Although these techniques vary in their mechanisms, they commonly require additional post hoc optimization separate from the classifier.

\subsection{Deep Interpretable Models}

Deep interpretable models have emerged as a promising avenue to reconcile high predictive performance with intrinsic explainability in tabular data. The early work of Alvarez-Melis and Jaakkola \citep{alvarez2018towards} pioneered the idea of self‐explaining neural networks that enforce local linearity and stability, laying the groundwork for models that can be interpretable by design. Based on these concepts, attention‐based architectures such as TabNet \citep{Arik2021tabnet} have demonstrated that sequential attention mechanisms can dynamically select the most relevant features during decision-making, offering built-in interpretability alongside competitive accuracy. Later, architectures like the Interpretable Mesomorphic Neural Networks (IMN) \citep{kadra2024interpretable} leverage deep hypernetworks to generate instance‐specific linear models, providing direct feature attributions without compromising the expressive power of deep networks. Complementary approaches include Neural Additive Models (NAMs) \citep{agarwal2021nams}, which combine the flexibility of deep learning with the transparency of generalized additive models through per-feature shape functions, and Deep Abstract Networks (DANet) \citep{chen2022danets}, which utilize specialized abstract layers to hierarchically group and transform features for clearer semantics. 

Collectively, these advances mark a significant evolution in deep interpretable modeling, offering unified frameworks where prediction accuracy and built-in explainability are integrated from the ground up, but none of these methods inherently provides counterfactual explanations, which remain an external, post-hoc addition.

\section{Preliminaries}
%%%%%%%%%%%%%%%%%%%%%%%%%%%%%%%%%%%%

\begin{table}[t]
\centering
\caption{Main notation used throughout the paper.}
\footnotesize
\label{table:notation}
\begin{tabular}{lcl}
\toprule
 & \textbf{Symbol} & \textbf{Description} \\
\midrule
\multirow{3}{*}{\rotatebox{90}{\textbf{Dims}}} 
& $D$ & Number of input features \\
& $K$ & Number of classes \\
& $N$ & Number of training samples \\
\midrule
\multirow{11}{*}{\rotatebox{90}{\textbf{Variables}}} 
& $x \in \R^D$ & Input data point \\
& $x' \in \R^D$ & Counterfactual example \\
& $y \in \{1,\ldots,K\}$ & True class label \\
& $k \in \{1,\ldots,K\}$ & Predicted class label \\
& $m \in \{1,\ldots,K\}$ & Target class for counterfactual \\
& $W \in \R^{K \times (D+1)}$ & Weight matrix output by hypernetwork \\
& $W_m$ & $m$-th row of weight matrix $W$ \\
& $z(x;W) \in \R^K$ & Logits vector for $x$ determined by $W$\\
& $z(x;W)_k$ & Logit of the $k$-th class \\
& $\theta$ & Parameters of hypernetwork $H$ \\
& $\phi$ & Parameters of normalizing flow $F$ \\
\midrule
\multirow{10}{*}{\rotatebox{90}{\textbf{Functions \& Notations}}} 
& $f: \R^D \to \R^K$ & Classifier function \\
& $H(\cdot;\theta): \R^D \to \R^{K \times (D+1)}$ & Hypernetwork with parameters $\theta$ \\

& $F(\cdot|\cdot;\phi)$ & Conditional normalizing flow with parameters $\phi$ \\
& $\lcal_{CE}$ & Cross-entropy loss \\
& $\lcal_{conEx}$ & Counterfactual loss \\
& $\lcal_F$ & Flow plausibility loss \\
& $d(\cdot, \cdot)$ & Distance measure \\
& $p_F(x|y)$ & Conditional probability density \\
& $\alpha_1, \alpha_2, \alpha_3$ & Loss regularization hyperparameters \\
& $\delta$ & Density threshold for plausibility \\
\bottomrule
\end{tabular}
\end{table}

To describe \our{}, we first introduce the basic concept of counterfactual explanations. Next, we present two main components of \our{}: hypernetwork \citep{ha2016hypernetworks} and invertible normalizing flows \citep{rezende2015variational}. The key symbols used throughout the paper are summarized in Table~\ref{table:notation}.  

\begin{figure*}[!ht]
    \centering
    \includegraphics[width=\linewidth]{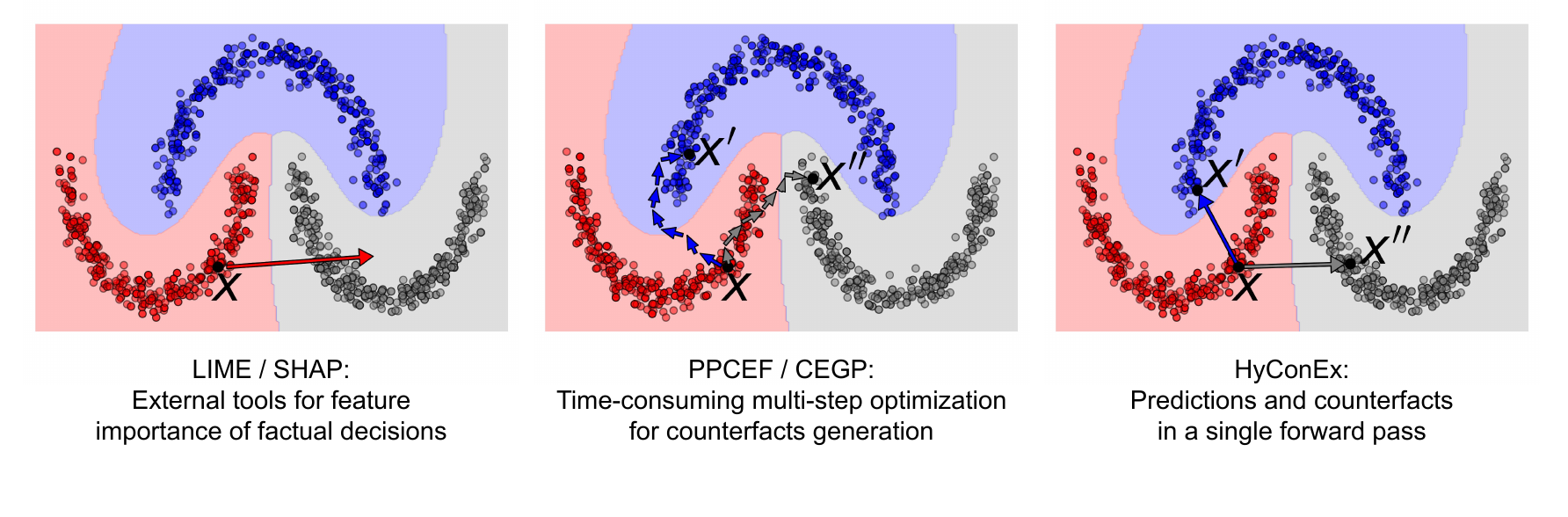}
    \caption{Comparison of \our{} with related explanability methods. Feature attribution methods, like LIME or SHAP, deliver static explanations supporting only the current decision. External methods for generating counterfactual examples, e.g. PPCEF or CEGP, typically perform time-consuming optimization for every input data point. \our{} encloses predictions and counterfactuals in a single forward pass, which makes it an extremely efficient method that delivers a dynamic explanation.}
    \label{fig:teaser}
\end{figure*}

\subsection{Counterfactual explanations}

Counterfactuals aim to explain classifier decisions by examples~\citep{verma2020counterfactual}. They indicate the modifications of the input examples, which are required to flip the classifier decision. In practice, counterfactual examples represent samples generated from the alternative (non-factual) classes, which are similar to the input.
By analyzing the difference between input and output examples, we obtain counterfactual explanations.  

To describe a basic approach to generating counterfactual examples, let ${f:\R^D \to \R^K}$ be a multi-class classifier trained on a given dataset $\{(x_i,y_i)\}_{i=1}^N$, where $D$ denotes the number of input features, $K$ is the number of classes and $N$ is the number of training data. We assume that the classifier $f$ returns a class label $k$ for a data point $x$. The question is: \emph{how to modify $x$ to obtain an alternative class $m$?} Since we are interested in local decisions, we want to make as small changes to $x$ as possible. Formally, counterfactual example $x'$ can be defined as a solution to the optimization problem \citep{wachter2017counterfactual}: 
\begin{equation}
\mathrm{arg min}_{ x' \in \R^D}
\lcal(f(x'), m) + C \cdot d(x, x'),
\end{equation}
where $\lcal(\cdot, \cdot)$ represents a loss function for classification, $d(\cdot, \cdot)$ is a distance function (proximity measure) acting as a penalty for deviations from the original input $x$, and $C > 0$ indicates the hyperparameter for regularization intensity.

Contemporary methods satisfy many additional properties, such as sparsity, actionability, or diversity, which increase their interpretation capabilities~\citep{verma2020counterfactual}. In this paper, we are especially interested in constructing plausible counterfactuals~\citep{Guidotti2024counterfactual}. A counterfactual is plausible if it is located within the high-density region of data. This feature distinguishes counterfactuals from adversarial examples. Plausibility is typically assessed via a straightforward k-neighborhood analysis of the counterfactual relative to the original dataset \citep{augustin2022diffusion}. Recently, PPCEF~\citep{Wielopolski2024ProbabilisticallyPC} has employed invertible normalizing flows to evaluate plausibility in tabular data contexts. Our study adopts a similar approach by integrating invertible normalizing flows directly into the classification model. A brief comparison of \our{} with other methods delivering counterfactuals and decision explanations is presented in Figure~\ref{fig:teaser}.

\subsection{Hypernetwork classifiers} \label{sec:hyper}
Hypernetworks, introduced in \citep{ha2016hypernetworks}, are defined as neural models that generate weights for a separate target network that solves a specific task. %In this seminal work, the authors aim to reduce the number of trainable parameters by designing a hypernetwork with fewer parameters than the target network. The hypernetworks were developed and applied in several areas including generative modeling, continual learning \citep{von2020continual}, image representations \citep{klocek2019hypernetwork}, and many others. 
% Making an analogy between hypernetworks and generative models, the authors of \citep{sheikh2017stochastic}, use this mechanism to generate a diverse set of target networks approximating the same function.
The recently introduced IMN~\citep{kadra2024interpretable} uses a hypernetwork to construct an individual local linear classifier for every tabular data point. %Due to the linearity of the classifier, IMN also delivers the local importance of the input features (determined by the linear model). 
Although IMN uses local linear models, it produces a global non-linear decision boundary in the whole data space. 

% \przmek{Let us denote a tabular dataset consisting of $N$ instances of $D$--dimensional features as $X \in \R^{D \times M}$ and the categorical target variable as $Y \in \{1,\ldots,C\}^K$.}

In this approach, the hypernetwork ${H(\cdot;\theta): \R^D \to \R^{K \times (D+1)}}$, parameterized by weights $\theta$, takes a data point $x \in \R^D$ and outputs the local linear classifier for $K$-class problem given by the weight matrix $W = H(x;\theta)$. The $k$-th row of $W$ (denoted by $W_k$) determines the logit of the $k$-th class as follows:
\begin{equation}
z(x;W)_k = \sum_{d=1}^D W_{k,d} x_{d} + W_{k,0},
\end{equation}
where $z(x;W)_k$ is the $k$-th coordinate of the vector $z(x;W)$. In consequence, the probability that $x$ belongs to $k$-th class is parametrized by applying a softmax function to the logits $z(x,W)$:
\begin{equation}
f(x;W)_k = \frac{\exp({z(x;W)_k})}{ \sum_{j=1}^{K}\exp({z(x;W)_j})}.
\end{equation}
Since prediction for every data point is determined by possibly different linear model, the overall decision boundary on the data space can be highly non-linear. The hypernetwork is instantiated by Tabular ResNet \citep{kadra2024interpretable}.

\begin{figure*}[t]
    \centering
    \subcaptionbox{Pure cross-entropy for the input sample: no class transition.\label{fig:comp_a}}{\includegraphics[width=0.24\linewidth, trim=50 10 50 10, clip]{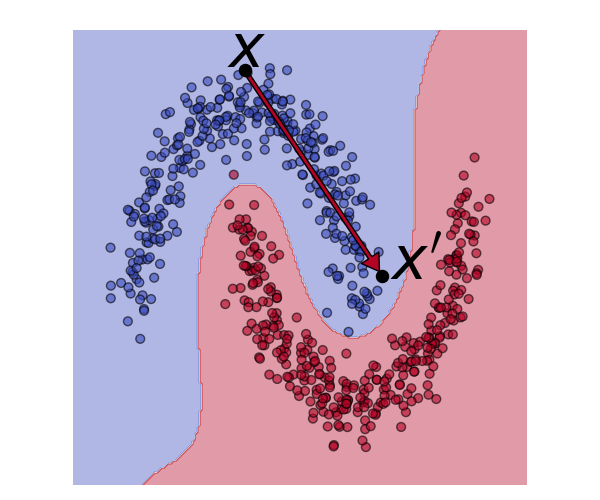}} %
    \subcaptionbox{Addition of the cross-entropy loss for the counterfact: class transition, out-of-distribution sample, long distance between the input and the counterfact.\label{fig:comp_b}}{\includegraphics[width=0.24\linewidth, trim=50 10 50 10, clip]{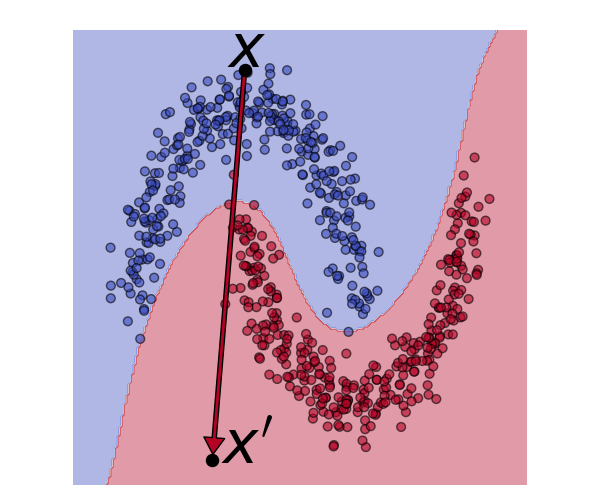}} %
    \subcaptionbox{Addition of the proximity loss: minimization of the distance between the input and the counterfact, class transition, out-of-distribution counterfact.\label{fig:comp_c}}{\includegraphics[width=0.24\linewidth, trim=50 10 50 10, clip]{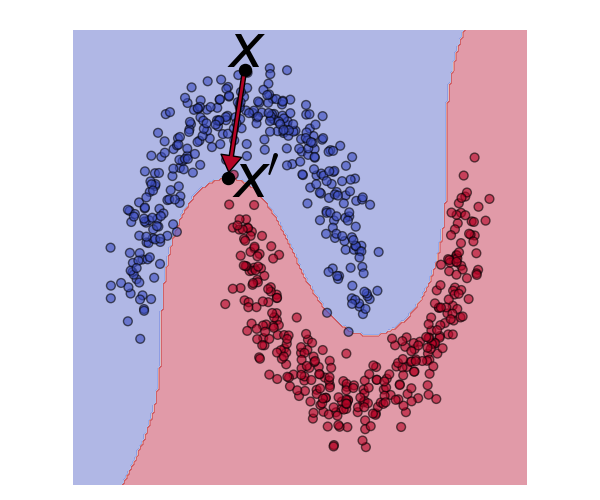}} %
    \subcaptionbox{Addition of the plausibility loss: ensuring the in-distributed counterfactual sample.\label{fig:comp_d}}{\includegraphics[width=0.24\linewidth, trim=50 10 50 10, clip]{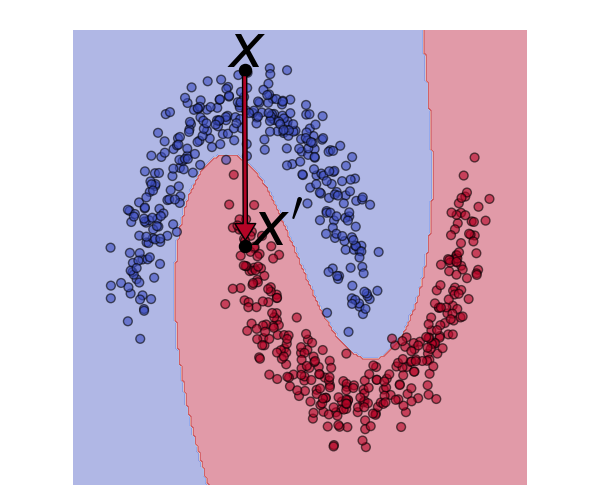}} %
    \caption{Illustration of the impact of all components of our cost function: (a) Cross-entropy is applied only to the input example similar to  IMN \citep{kadra2024interpretable}. As we can see, the normal vector $w$ is turned into the target class, but the translation of $x'=x-w$ does not lead to the change of class label; (b) We additionally apply the cross-entropy to the counterfactual example. In this case, we produce a strong modification, which does not lead to in-distribution samples; (c) We supply the previous loss with the proximity between input and counterfactual. We observe that the model gives the smallest modification, which changes the class label. (d) Complete loss with plausibility term. The final \our{} enforces $x'$ to lie in the manifold of the target class.
    }
    \label{fig:comp}
\end{figure*}

Let us observe that the linear model determined by ${W = H(x;\theta)}$ contains the local normal vectors to the linear classification boundary for the input $x$. In other words, it gives a direction to the decision boundary, see Figure~\ref{fig:comp_a}, which can be understood as \emph{feature importance} vector for $x$. Unfortunately, these vectors represent a chosen class against all other outcomes. Our study focuses on generating counterfactual examples in a single forward pass, which provides more profound insights.

\subsection{Invertible Normalizing Flows}

To generate plausible counterfactuals, we need to model the density of data~\citep{rezende2015variational}. This will be 
estimated using invertible normalizing flows (INFs).

INFs can be seen as sophisticated tools for estimating data density. In practice, INFs integrate autoencoder structures where the encoder is an invertible function. Due to these characteristics, the inverse function can serve as a decoder. These models do not reduce dimensionality, making them generally suitable for representing relatively low- or medium-dimensional data. This makes them particularly well-suited for handling tabular data.
INFs are trained using negative log-likelihood (NLL) as a cost function. 

INFs transform a latent variable $z$ with a known prior distribution $p(z)$ into an observed space variable $x$ with an unknown distribution. This mapping is achieved through a sequence of invertible parametric transformations:
\begin{equation}
 x = F(z) = g_L \circ \ldots \circ  g_1(z),
\end{equation} where each function $g_l$, for $l=1,\ldots,L$, progressively transforms the latent representation toward the observed data space.

To model a separate density for each class, INF is additionally conditioned by a class vector $y$. In this case, the conditional log-likelihood for $x$ is expressed as:
\begin{equation}
\log p_F (x|y) = \log p(z) - \sum^{L}_{l=1}
\log \bigg|
det \frac{\partial g_l }{\partial  z_{l-1}}
\bigg|,
\end{equation}
where $z = z_0 = g^{-1}_1 \circ \ldots \circ  g^{-1}_L (x, y)$ is the latent obtained by inverting the flow, and $z_l = g_l(z_{l-1})$ denotes the intermediate flow variable at layer $l$.
A significant challenge in INFs is the selection of appropriate invertible functions, The existing literature presents a variety of solutions to address this issue, with notable methodologies including Non-linear Independent Components Estimation (NICE)~\citep{dinh2014nice}, Real-Valued Non-Volume Preserving (RealNVP)~\citep{dinh2016density}, and Masked Autoregressive Flow (MAF)~\citep{papamakarios2017masked}.

\section{\our{} model}
%%%%%%%%%%%%%%%%%%%%%%%%%%%%%%%%%%%%

This section describes \our{}, which is a deep learning classifier equipped with counterfactual explanations. Since \our{} is an all-in-one neural network,  predictions and counterfactuals are returned in a single forward pass. Table~\ref{table:notation} summarizes the key notation used in our model description.

\subsection{Architecture}
\label{sec:architecture}

\begin{figure*}[t]
    \centering
    \includegraphics[width=0.95\linewidth]{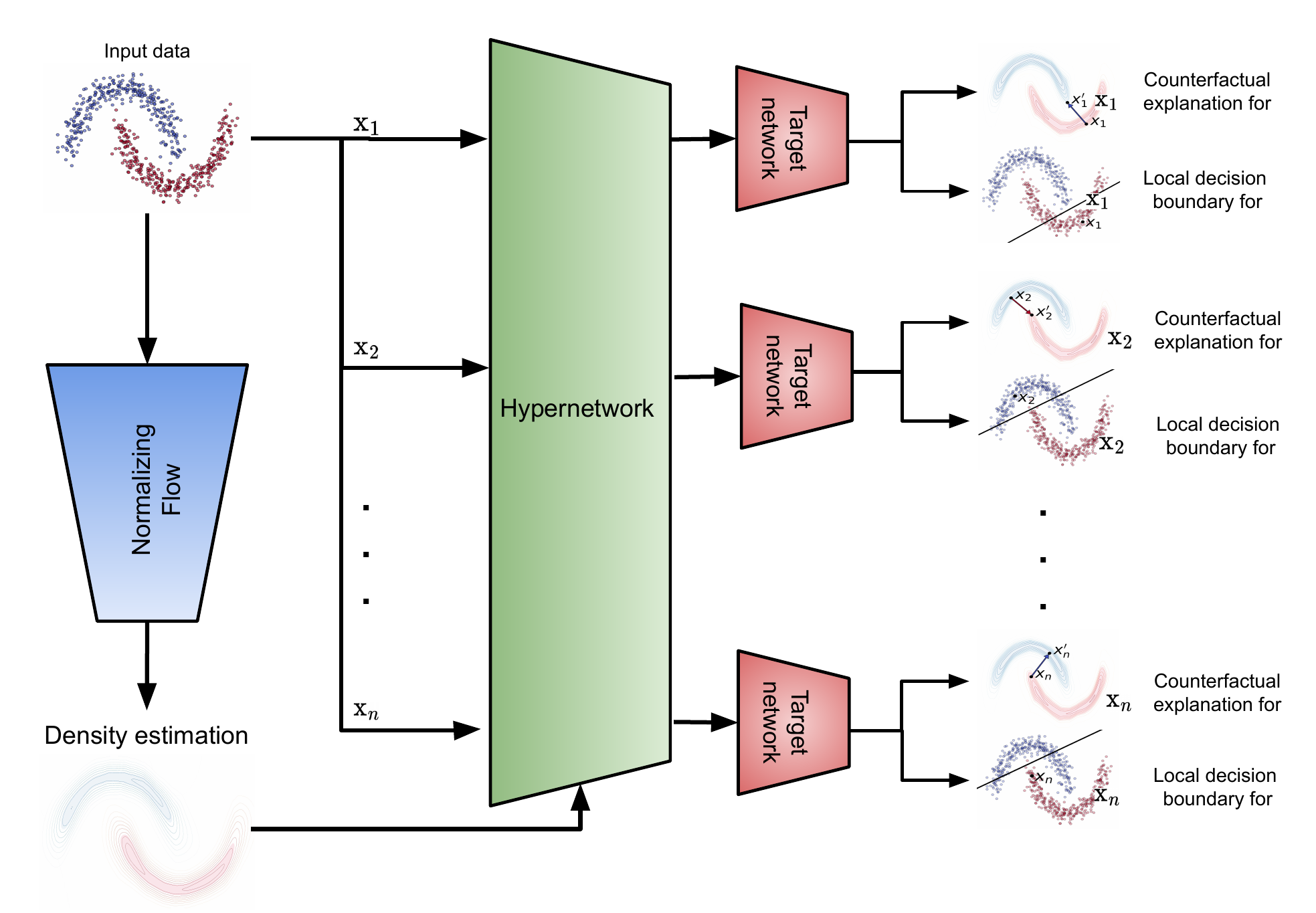}
    \caption{Illustration of \our{} architecture. The hypernetwork takes every input together with the estimated data density and returns: (1) a local decision boundary, and (2) counterfactual examples for all alternative classes. In consequence, we obtain predictions and counterfactuals in a single forward pass.}
    \label{fig:arch}
\end{figure*}

Our architecture comprises two key elements: hypernetwork classifier and invertible normalizing flows, see Figure~\ref{fig:arch}. By defining a proper loss function and designing a dedicated training procedure, the generated classifier delivers predictions and the associated counterfactual explanations.

Let us construct a hypernetwork $H(\cdot;\theta)$, which returns a weight matrix ${W=H(x;\theta)}$ that defines a local linear classifier $f(\cdot;W)$ as described in Section~\ref{sec:hyper}. We will force the hypernetwork to produce such a weight matrix $W$, which not only contains information needed for class predictions but also gives the localization of counterfactual examples for alternative classes. To keep our model as simple as possible, we do not introduce any additional parameters. The counterfactual examples will be determined by the rows of weight matrix $W$.

To explain our approach, we assume that $x \in \R^D$ is classified by $f(x;W)$ as the element of the class $k$. 
It means that the $k$-th row of $W$ is related to the vector that is normal to the local decision boundary generated for~$x$. To localize the counterfactual examples for any other class~$m$, we utilize the remaining rows of $W$, denoted as $W_m$. Concretely, we define a counterfactual candidate $x'$ as a translated version of x, given by $x' = x - W_m$. For $x'$ to
indeed represent a valid counterfactual instance of class~$m$, the following conditions must be met:
\begin{enumerate}
    \item The classifier $f(\cdot; H(x'; \theta))$ assigns $x'$ to class $m$,
    \item $x'$ is close to the original input $x$,  
    \item $x'$ is plausible (in-distribution sample). 
\end{enumerate}

The first criterion is satisfied by minimizing the cross-entropy loss $\lcal_{CE}(f(x'; W'), m)$, where $W' = H(x'; \theta)$ is the weight matrix returned by the hypernetwork~$H$ for instance~$x'$ (as shown in Figure~\ref{fig:comp_b}). Second, to satisfy the proximity between $x$ and $x'$, we minimize the MSE between them, i.e. $\|x-x'\|^2$, see Figure~\ref{fig:comp_c}. And the final one is addressed using class-conditional
invertible normalizing flow $F(\cdot|m; \phi)$, with parameters $\phi$, which estimates the data density within each class separately (as illustrated in Figure~\ref{fig:comp_d}). Following the approach proposed in PPCEF~\citep{Wielopolski2024ProbabilisticallyPC}, we employ a cost function that encourages $x'$ to reach a fixed density threshold. This function is given by:
\begin{equation}
\lcal_F(x',m; \phi) = \max\left(\delta - p_F(x'|m; \phi),0\right),
\end{equation}
where $p_F(x'|m)$ represents the probability density of the counterfactual example $x'$, conditioned on class $m$, and $\delta$ is a threshold set based on the median density of the training data.

By combining these three cost components into a single expression, we obtain the
counterfactual loss $\lcal_{conEx}$ for a given alternative class $m$:
\begin{equation}
\lcal_{conEx}(x,x',m;\theta,\phi) =
\alpha_1 \lcal_{CE}(f(x';H(x';\theta)), m) + \alpha_2 \| x - x'\| + \alpha_3 L_F(x',m; \phi),
\end{equation}
where $\alpha_1$, $\alpha_2$, and $\alpha_3$ are regularization hyperparameters that balance the contributions of the three respective terms.

In the complete loss function, we incorporate counterfactual losses $\lcal_{conEx}$ for every
target class, as our goal is to enable the generation of counterfactuals in all possible directions.
Additionally, we include the standard cross-entropy component $\lcal_{CE}(f(x; W), y)$ to ensure
correct classification, where $y$ denotes the true label of $x$. Finally, the full loss $\lcal_{\our}$ is
defined as:

\begin{equation} \label{eq:loss}
\lcal_{\our}(x, x', y;\theta, \phi) =
\lcal_{CE}(f(x; W), y) + \sum_{m \neq y} \lcal_{conEx}(x,x',m;\theta,\phi).
\end{equation}

\subsection{Initialization and training process of \our{}}

The entire training process of \our{} consists of three phases: pre-training of the hypernetwork \( H(\cdot;\theta) \), training of the conditional INF \( F(\cdot; \phi) \), and fine-tuning of the hypernetwork using Equation~\ref{eq:loss}. All phases are described below:
\begin{enumerate}
    \item Since the hypernetwork produces not only a classification boundary but also counterfactual explanations, we need to pre-train \our{} in a way that ensures a suitable starting point for the optimization process detailed in Section~\ref{sec:architecture}. To this end, we use clustering with the k-means algorithm \citep{mcqueen1967some}, which partitions each class into several components. The centers of the obtained regions provide a reasonable approximation of counterfactual examples. During this phase, we utilize the classification cross-entropy loss $\lcal_{CE}(f(x), y)$, combined with the proximity loss between counterfactual candidate $x'$ and the nearest clustering mean of the alternative class:
$$
 \lcal_{CE}(f(x), y) + \alpha \sum_{m \neq y} \| x' - r_m\|,
$$
where $x' = x - W_m$ and $r_m$ is the  closest cluster center from the $m$-th class to a data point $x$ and $\alpha$ is a hyperparameter. This clustering-based pre-training is targeted to obtain plausible counterfactuals, but with lower quality than applying INFs. However, it is more stable, which is important at the initial stage of the training.

    \item In the second stage, we train a conditional INF $F(\cdot; \phi)$ on class labels predicted by a pre-trained hypernetwork classifier.
This initialization helps to stabilize the training process and leads to faster convergence. 
\item After pre-training stages, we freeze the parameters of INF and train only the hypernetwork network until convergence, see Algorithm~\ref{alg:tr}. The inference is presented in Algorithm~\ref{alg:inf} for completeness.
\end{enumerate}

\begin{algorithm}[H]
    \caption{\our{} Training Iteration}\label{alg:tr}
    \begin{algorithmic}
    \State \textbf{Input:} Hypernetwork $H(x; \theta)$ returning a linear model $f(x;W)$ by $W = H(x;\theta)$, \\
    Conditional normalizing flow $F(\cdot |y;\phi)$ for class $y$,\\ 
    Training set $\mathcal{D}$ and set of class labels $\mathcal{C}$
    \For{each batch $(x, y)$ in $\mathcal{D}$}
        \State $W \gets H(x;\theta)$
        \Comment{Linear model}
        \State $\hat{y} \gets f(x;W)$ \Comment{Current predictions for $x$}
        \State $loss \gets \lcal_{CE}(\hat{y}, y)$
        \For{each $m$ in $\mathcal{C}$}
            \State $x' \gets x - W_m$ \Comment{Counterfactual for $x$ and class $m$} 
            \If{$m \ne y$}
                \State $loss \mathrel{+}= \lcal_{conEx}(x, x', m; \theta, \phi)$
            \EndIf
        \EndFor
        \State Update $\theta$ with respect to $loss$ via gradient optimization
    \EndFor
    \end{algorithmic}
\end{algorithm}

\begin{algorithm}[H]
    \caption{\our{} Inference} \label{alg:inf}
    \begin{algorithmic}
    \State \textbf{Input:} Hypernetwork $H(x; \theta)$ returning a linear model $f(x;W)$ by $W = H(x;\theta)$,\\
    Factual instance $x$ and target class $t$
    \State $W \gets H(x;\theta)$
        \Comment{Linear model}
        \State $\hat{y} \gets f(x;W)$ \Comment{Current predictions for $x$}
        \State $x' \gets x - W_t$ \Comment{Counterfactual for $x$ and class $t$}
    \State\Return $\hat{y}, x'$ 
    \end{algorithmic}
\end{algorithm}

\subsection{Categorical features} \label{sec:cat}

To represent categorical variables numerically, we use one-hot encoding. After applying this transformation, we could theoretically assume that we are now dealing with a fully numerical problem and expect the generated counterfactual vectors to be close to one-hot vectors in the categorical positions, because of the INF component in the loss function. However, this approach unfortunately fails and leads to validity issues, as demonstrated by the PPCEF results in Table~\ref{table:countefacts_categorical}. 

Therefore, to address this problem, we additionally introduce special smoothing on the categorical positions using softmax with a low temperature $T$. More precisely, we modify the generated vector $x'$ at positions from $i_1$ to $i_K$, which correspond to the one-hot encoded range of a given categorical variable, as follows:
$$
    x'_{i_k} = \frac{\exp\left(\frac{x'_{i_k}}{T}\right)}
    { \sum_{j=1}^{K} \exp\left(\frac{x'_{i_j}}{T}\right)}.
$$
Lower values of $T$ lead to the formation of more one-hot-like vectors. In our experiments, we found that setting $T=0.01$ produced stable and satisfactory results. Therefore, we do not treat $T$ as a tunable hyperparameter but instead fixed it at this value for all experiments.

The entire process is repeated separately for each categorical feature in the dataset. Implementing this idea allows us to produce counterfactuals with near-perfect validity without compromising metrics related to density or proximity.

\section{Experiments}

We evaluate \our{} in terms of classification accuracy and counterfactual explanation quality. To our knowledge, \our{} is the first deep learning model that returns counterfactual examples by a classification model itself. %Since this is one of the contributions of our paper, we will evaluate it in more detail.

% Since \our{} can be seen as the extension of IMN, we expect that predictive accuracy remains at the analogical level and returned importance of features give similar information to the user. The counterfactual explanation delivered by \our{} will be compared with recent and well-established baselines in this area.

\subsection{Instantiation of \our{}}

\begin{table}[t]
\footnotesize
\centering
\caption{Hyperparameters of \our{}.}
\label{table:hyperparmas}
\begin{tabular}{lcll}
\toprule
 & \textbf{Name} & \textbf{Description} & \textbf{Default value} \\
\midrule
\multirow{4}{*}{\rotatebox{90}{\textbf{Backbone}}} 
& \textit{nr\_blocks} & Number of residual blocks & 4 \\
& \textit{hidden\_size} & Hidden layer size & 256 \\
& \textit{dropout\_rate} & Dropout probability & 0.25 \\
& \textit{act\_func} & Activation function & GeLU \\
\midrule
\multirow{5}{*}{\rotatebox{90}{\textbf{MAF}}} 
& \textit{num\_layers} & Number of autoregressive layers & 8\\
& \textit{hidden\_features} & Hidden dimension size & 16\\
& \textit{num\_blocks\_per\_layer} & Blocks per autoregressive layer & 4 \\
& \textit{dropout\_rate} & Dropout probability & 0.00 \\
& \textit{act\_func} & Activation function & ReLU \\
\midrule
\multirow{13}{*}{\rotatebox{90}{\textbf{Training}}}
& \textit{nr\_epochs} & Maximal number of train epochs & 1500 \\
& \textit{pretraining\_epochs} & Number of pretraining epochs & 500 \\
& \textit{warm\_up\_epochs} & Warm-up epochs for $\lcal_{conEx}$ & 200 \\
& & components scaling & \\
& \textit{batch\_size} & Dataloader batch size & 256\\
& \textit{learning\_rate} & Learning rate & 5e-4 \\
& \textit{optimizer} & Optimization method & Adam \\
& \textit{scheduler} & LR scheduler & LambdaLR + \\
& & & CosineAnnealing \\
& $\alpha$ &  Initialization phase trade-off & 0.8\\
& $\alpha_1$ & $\lcal_{CE}$ loss weight & 0.8 \\
& $\alpha_2$ & MSE loss weight & 0.1\\
& $\alpha_3$ & $\lcal_F$ loss weight & 0.1\\
\bottomrule
\end{tabular}
\end{table}

For implementing \our{}, we use TabResNet as a hypernetwork backbone and MAF as conditional INF. TabResNet uses 4 residual blocks, a hidden size of 256, and a dropout rate of 0.25. MAF is configured with 16 hidden features, 8 layers, and 4 blocks per layer. 

The initialization phase uses a trade-off parameter $\alpha = 0.8$ and is continued until model convergence. In the second phase, we employ the same Adam optimizer and cosine annealing scheduler as used during pre-training. For all experiments, we linearly increase the trade-off parameters $\alpha_i$ from 0 to $\alpha_1$=0.8, $\alpha_2$=0.1 and $\alpha_3$=0.1. The duration of the second phase is determined using early stopping, which aims to maximize plausibility while ensuring high model accuracy, high counterfactual validity, and low proximity. 

All details are described in Table~\ref{table:hyperparmas}.

\subsection{Predictive performance}

To validate the predictive performance of \our{}, we use the benchmark from~\citep{kadra2024interpretable} with an identical experimental setup. In particular, we compare \our{} with strong ensemble methods (Random Forest and CatBoost), classical shallow classifiers (decision trees, logistic regression), as well as deep learning models equipped with interpretability tools (TabNet~\citep{Arik2021tabnet}, TabResNet, IMN~\citep{agarwal2021nams}). Our main goal is to show that introducing counterfactual functionality to a hypernetwork classifier does not deteriorate its original performance. Hence, we are particularly focused on comparing with IMN. 

We use 18 tabular datasets retrieved from the OpenML repository of varied characteristics (see Table~\ref{table:test_performances} for datasets IDs). We split every dataset into train, validation and test in proportion: 60-20-20. For all experiments, we perform standard preprocessing: normalizing features and handling class imbalance via downsampling. We report the Area Under the Receiver Operating Characteristic curve (AUROC), which is well-suited for imbalanced data as well.

\begin{table*}[t]
\setlength{\tabcolsep}{1,5pt}
\centering
\caption{Comparison of predictive performance using AUROC measure. Introducing counterfactual explanations to hypernetwork classifier does not deteriorate its predictive performance (compare \our{} with IMN).}
\label{table:test_performances}
\centering
\footnotesize
\begin{tabular}{lccccccccc}
\toprule
Dataset & Dataset & Decision & Logistic & Random & TabNet & TabResNet & CatBoost & IMN & \our{} \\
ID & Name & Tree & Regression & Forest &  &  &  &  &  \\
\midrule
3 & kr-vs-kp & 0.987 & 0.990 & 0.998 & 0.983 & \textbf{0.999} & \textbf{0.999} & \textbf{0.999} & 0.997\\
12 & mfeat-factors & 0.938 & \textbf{0.999} & 0.998 & 0.995 & \textbf{0.999} & \textbf{0.999} & \textbf{0.999} & \textbf{0.999}\\
31 & credit-g & 0.643 & 0.775  & 0.795 & 0.511 & 0.756 & 0.790 & 0.751 & \textbf{0.799}\\
54 & vehicle & 0.804 & 0.938  & 0.927 & 0.501 & \textbf{0.968} & 0.934 & 0.957 & 0.946\\
1067 & kc1 & 0.620 & 0.802 & 0.801 & 0.789 & 0.808 & 0.800 & 0.805 & \textbf{0.809} \\
1461 & bank-marketing & 0.703 & 0.908  & 0.930 & 0.926 & 0.931 & \textbf{0.937} & 0.930 & 0.905 \\
1464 & blood-transfusion & 0.599 & 0.749 & 0.666 & 0.516 & 0.740 & 0.709 & 0.742 & \textbf{0.769} \\
1468 & cnae-9 & 0.926 & \textbf{0.996} & 0.995 & 0.495 & 0.995 & \textbf{0.996} & 0.994 & \textbf{0.996} \\
1486 & nomao & 0.935 & 0.987 & 0.993 & 0.991 & \textbf{0.994} & \textbf{0.994} & 0.993 & 0.987 \\
1489 & phoneme & 0.842 & 0.805 & \textbf{0.962} & 0.928 & 0.949 & 0.948 & 0.950 & 0.900\\
1590 & adult & 0.752 & 0.903 & 0.917 & 0.908 & 0.915 & \textbf{0.930} & 0.915 & 0.872 \\
40984 & segment & 0.946 & 0.980 & \textbf{0.995} & 0.985 & 0.994 & \textbf{0.995} & 0.994 & 0.985 \\
41142 & christine & 0.626 & 0.742 & 0.796 & 0.713 & 0.782 & \textbf{0.822} & 0.775 & 0.725 \\
41143 & jasmine & 0.749 & 0.850 & \textbf{0.880} & 0.823 & 0.860 & 0.870 & 0.865 & 0.835 \\
41146 & sylvine & 0.910 & 0.966 & 0.983 & 0.974 & 0.982 & \textbf{0.988} & 0.981 & 0.979 \\
41161 & riccardo & 0.857 & 0.995 & 0.999 & 0.997 & 0.998 & \textbf{1.000} & 0.998 & 0.994 \\
41163 & dilbert & 0.873 & 0.994 & 0.999 & 0.998 & \textbf{1.000} & \textbf{1.000} & \textbf{1.000} & 0.992 \\
41164 & fabert & 0.786 & 0.898 & 0.925 & 0.888 & 0.913 & \textbf{0.935} & 0.902 & 0.895 \\
% \midrule
% mean rank & 7.67 & 5.14 & 3.22 & 6.47 & 2.97 & 2.39 & 3.42 & 4.72 \\
% median rank & 8.00 & 5.50 & 2.75 & 7.00 & 3.00 & 2.00 & 3.50 & 5.75 \\
\bottomrule
\end{tabular}
\end{table*}

\begin{figure}[!ht]
    \centering
    \includegraphics[width=0.7\linewidth]{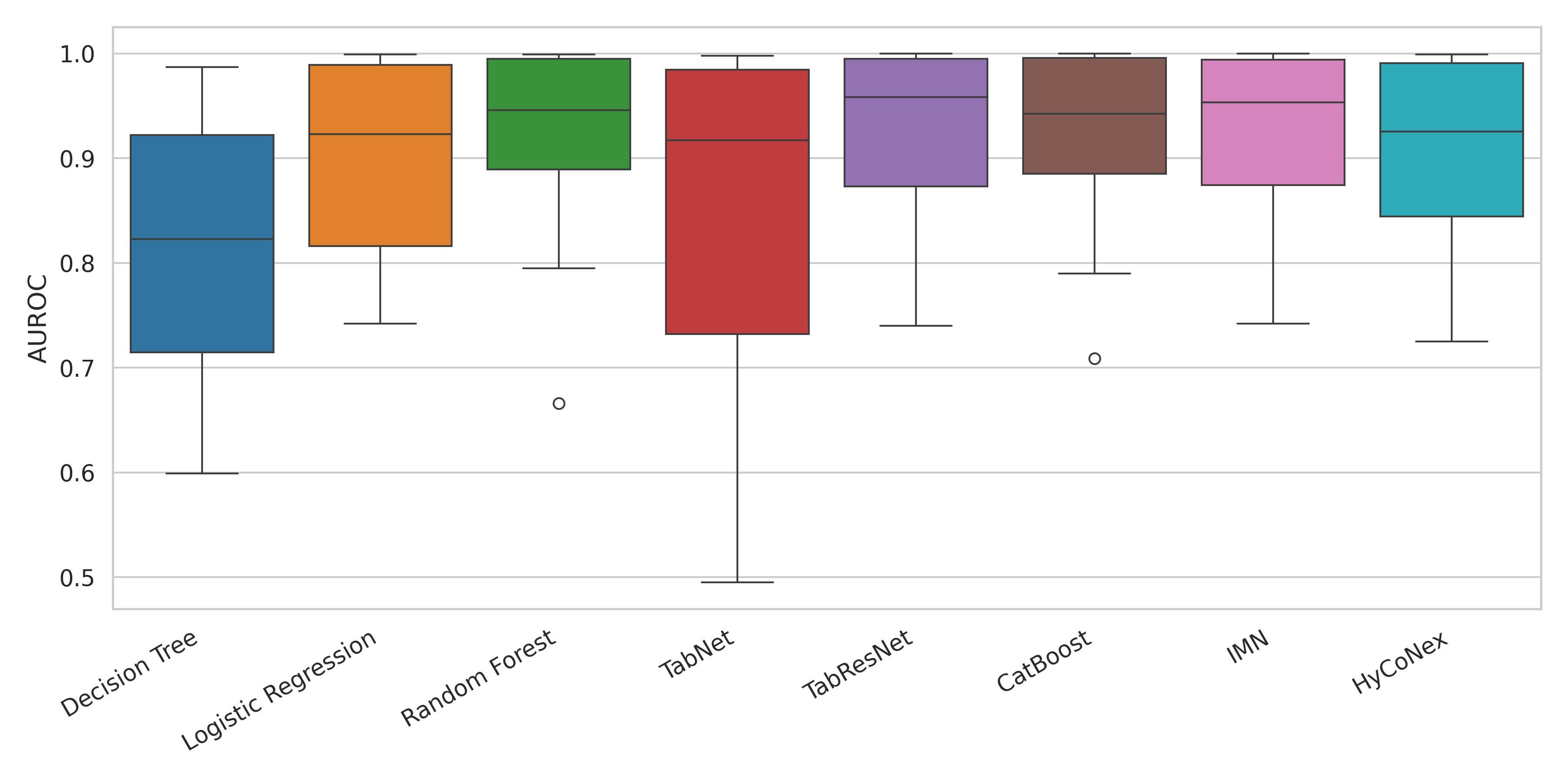}
    \caption{Mean AUROC scores of comparative methods.}
    \label{fig:auroc}
\end{figure}

The results presented in Table~\ref{table:test_performances} demonstrate that \our{} maintains overall strong performance, achieving the highest AUROC on five datasets (12, 31, 1067, 1464, 1468). Additionally, Figure~\ref{fig:auroc} shows that \our{} achieves fifth best mean AUROC score. These results reveal that additional counterfactual generation capability embedded in \our{} leads to a slight deterioration of classification performance. However, this degradation is not high.
% This confirms that our model effectively preserves predictive performance while gaining interpretability through counterfactual explanations. It is also evident that CatBoost outperforms other methods, which reaffirms that classical methods remain at the forefront of approaches for tabular data. However, these methods do not deliver any internal interpretations of their predictions.

To evaluate the significance of \our{}'s performance against competitive methods (measured by the AUROC values presented in Table~\ref{table:test_performances}), we performed a formal statistical comparison. We employed the one-sided Wilcoxon signed-rank test for paired samples and adjusted the obtained p-values using the Holm-Bonferroni correction to mitigate the risk of type I errors. As shown in Table~\ref{table:statistical_tests}, none of the benchmarked methods outperform \our{}. However, without adjusting the p-value, Random Forest, TabResNet, CatBoost and IMN would have suggested a significant performance advantage over \our{}.

\begin{table*}[!h]
\centering
\caption{Results of the Wilcoxon signed-rank test evaluating the \our{} performance measured by AUROC. Adjusted p-values represent the Holm-Bonferroni correction for multiple comparisons. Bold values correspond to p-values below the significance level threshold of $0.05$.}
\label{table:statistical_tests}
\footnotesize
\begin{tabular}{lrll}
\toprule
Method              & statistic & raw p-value & adjusted p-value \\ \midrule
Decision Tree       & 0.0 & 1.000                        & 1.000  \\
Logistic Regression & 45.0 & 0.803                        & 1.000  \\
Random Forest       & 124.5 & \textbf{0.047}                        & 0.223  \\
TabNet              & 44.5 & 0.935                        & 1.000  \\
TabResNet           & 122.5 & \textbf{0.015}                        & 0.103  \\
CatBoost            & 100.0 & \textbf{0.049}                        & 0.223  \\
IMN                 & 119.5 & \textbf{0.021}                        & 0.125 \\
\bottomrule
\end{tabular}
\end{table*}

\subsection{Counterfactual explanations}

\paragraph{Evaluation setup}

We evaluated counterfactual explanations by assessing the quality, plausibility, and computational efficiency of our \our{} method compared to established baselines such as CEM~\citep{dhurandhar2018cem}, CBCE~\citep{keane2020cbce}, CEGP~\citep{van2021cegp}, and PPCEF~\citep{Wielopolski2024ProbabilisticallyPC}. Our evaluation leverages a benchmark from  \citep{Wielopolski2024ProbabilisticallyPC} (with an identical setup) and expands it by incorporating additional categorical datasets.

We consider multiple scenarios, including binary classification using datasets such as \textit{Moons}, \textit{Law}, \textit{Audit} and \textit{Heloc}, multiclass classification using datasets such as \textit{Blobs}, \textit{Digits} and \textit{Wine}, and cases with categorical features using \textit{Adult} \citep{adult_2}, \textit{Credit-A} \citep{australian_credit_approval} and \textit{Credit-G} \citep{german_credit_data} datasets. For all experiments, we perform standard preprocessing: normalizing features and handling class imbalance via downsampling. For datasets with categorical features, we introduce slight Gaussian noise to labels for efficient training, while for counterfactual examples, we apply softmax activation to produce one-hot predictions and select the class with maximum value (see Section~\ref{sec:cat} for details).

Since \our{} is an interpretable model, which includes predictions and counterfactuals in one model, all methods generate counterfactuals for \our{} classifier. In this way, we provide a fair comparison of all methods. The test AUROC of \our{} for all datasets equals: Audit auroc=0.995, Credit-A auroc=0.932, Credit-G auroc=0.794, Moons auroc=1.000, Law auroc=0.834, Heloc auroc=0.798, Adult auroc=0.872, Wine auroc=1.000, Blobs auroc=1.000, Digits auroc=0.995.

\paragraph{Metrics}

We quantify counterfactual quality using several standard metrics from the counterfactual explanation literature \citep{Guidotti2024counterfactual}. Following the evaluation protocol of PPCEF \citep{Wielopolski2024ProbabilisticallyPC}, Coverage (Cover.) indicates the fraction of instances for which counterfactuals are generated, while Validity (Valid.) represents the percentage of counterfactuals that successfully changed the model prediction to the desired target. To assess the minimality of the changes from the original example, we report the L1 and L2 norms for continuous features and the Hamming distance for categorical features. Plausibility is measured using a probabilistic plausibility score (P.Plaus.) \citep{Wielopolski2024ProbabilisticallyPC}, which indicates how many counterfactuals have an estimated log density higher than the median log density of the train set. Furthermore, we provide additional plausibility metrics such as log density (LogDens) estimated using normalizing flow model and outlier detection scores such as Local Outlier Factor (LOF) \citep{breunig2000lof} and Isolation Forest (IsoForest) \citep{liu2008isolation}. LOF quantifies the local density deviation of a data point relative to its neighbors (higher values indicate outliers), while IsoForest measures how easily a point can be isolated in the feature space using binary trees (lower scores indicate outliers). Finally, we report the CPU computation time in seconds to generate counterfactuals to capture the efficiency of the method.

\begin{figure*}[t]
    \centering
\includegraphics[width=0.32\linewidth, trim=30 0 25 0, clip]{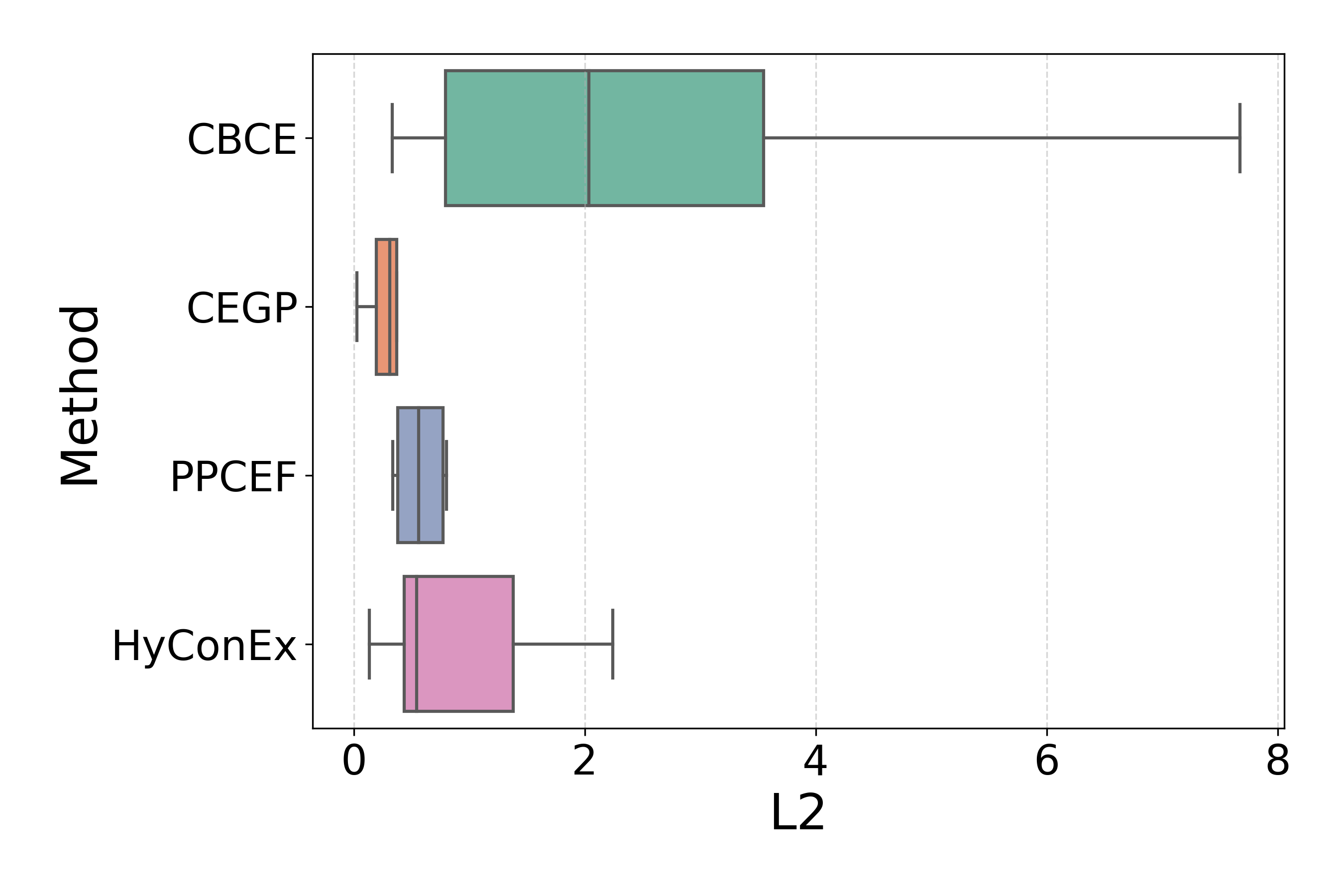}
\includegraphics[width=0.32\linewidth, trim=30 0 20 0, clip]{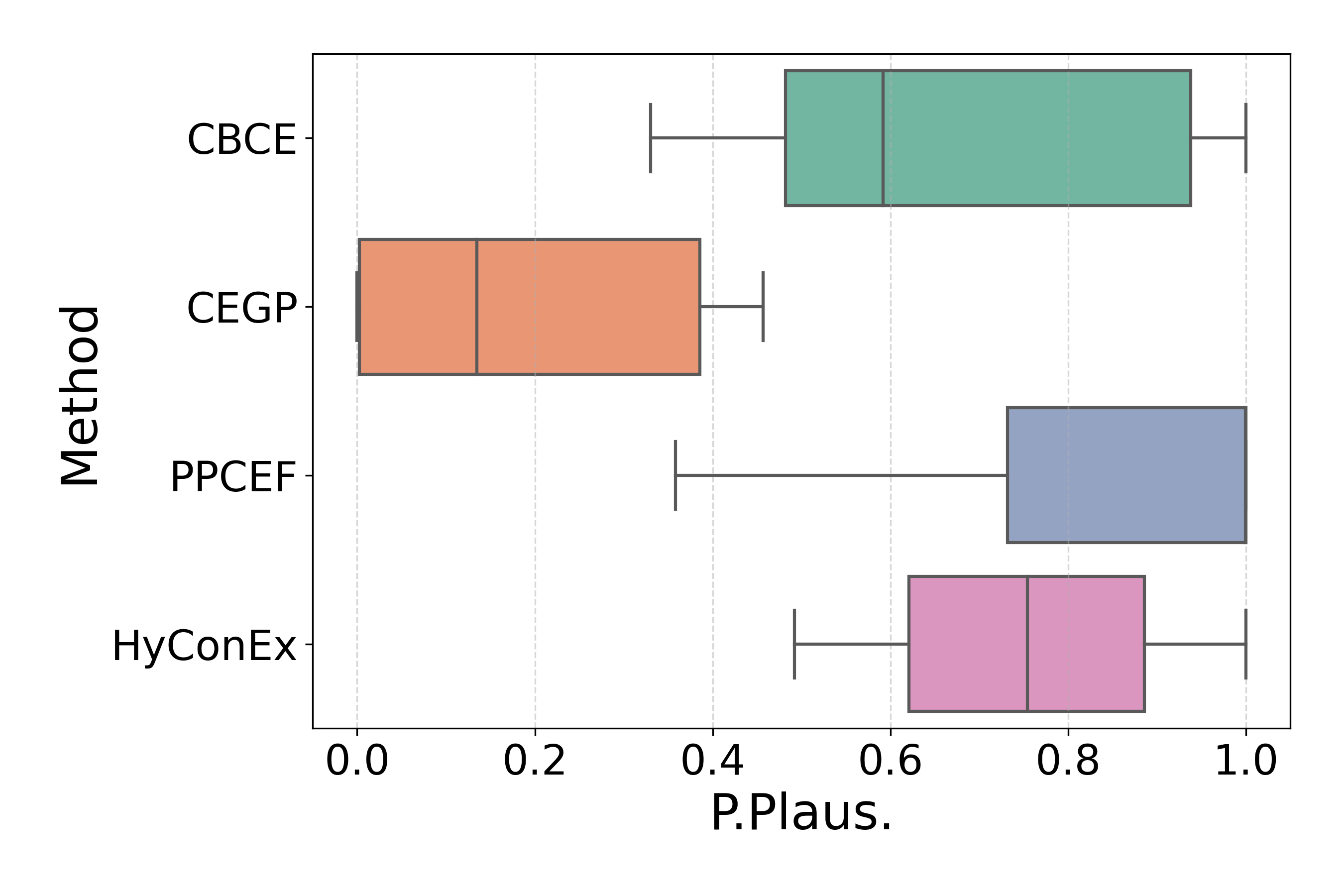}
\includegraphics[width=0.32\linewidth, trim=30 0 20 0, clip]{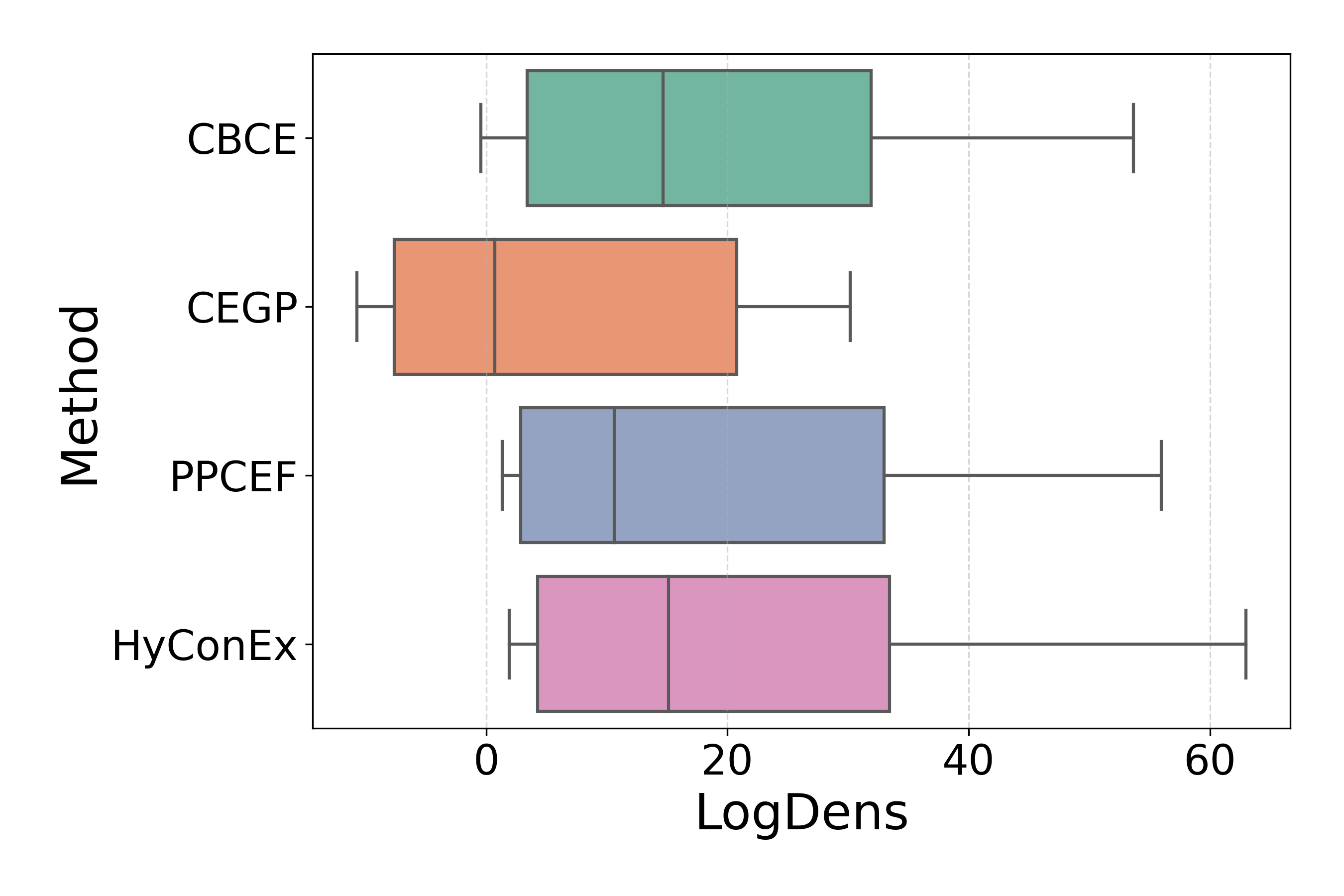}

\includegraphics[width=0.32\linewidth, trim=30 0 25 0, clip]{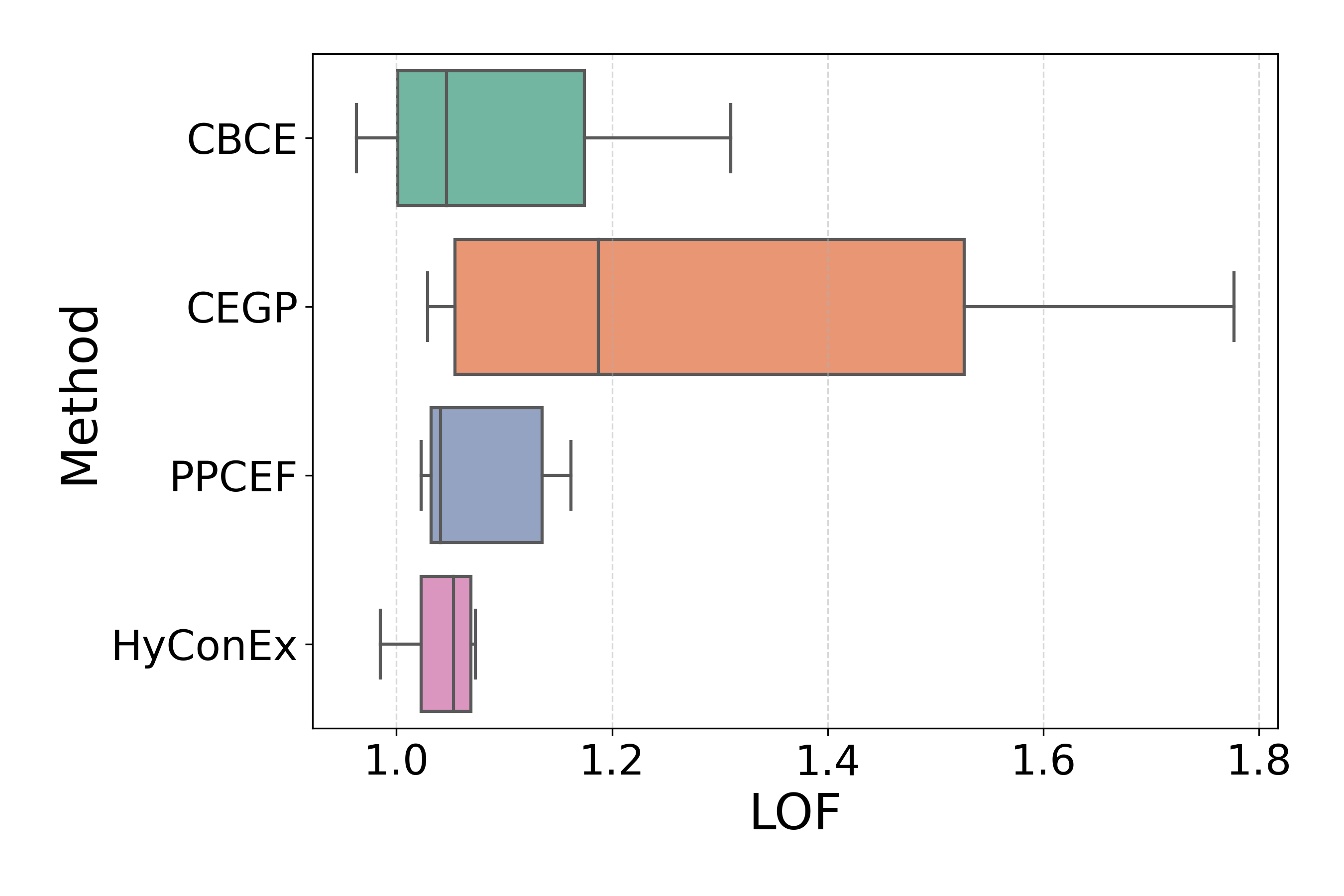}
\includegraphics[width=0.32\linewidth, trim=30 0 27 0, clip]{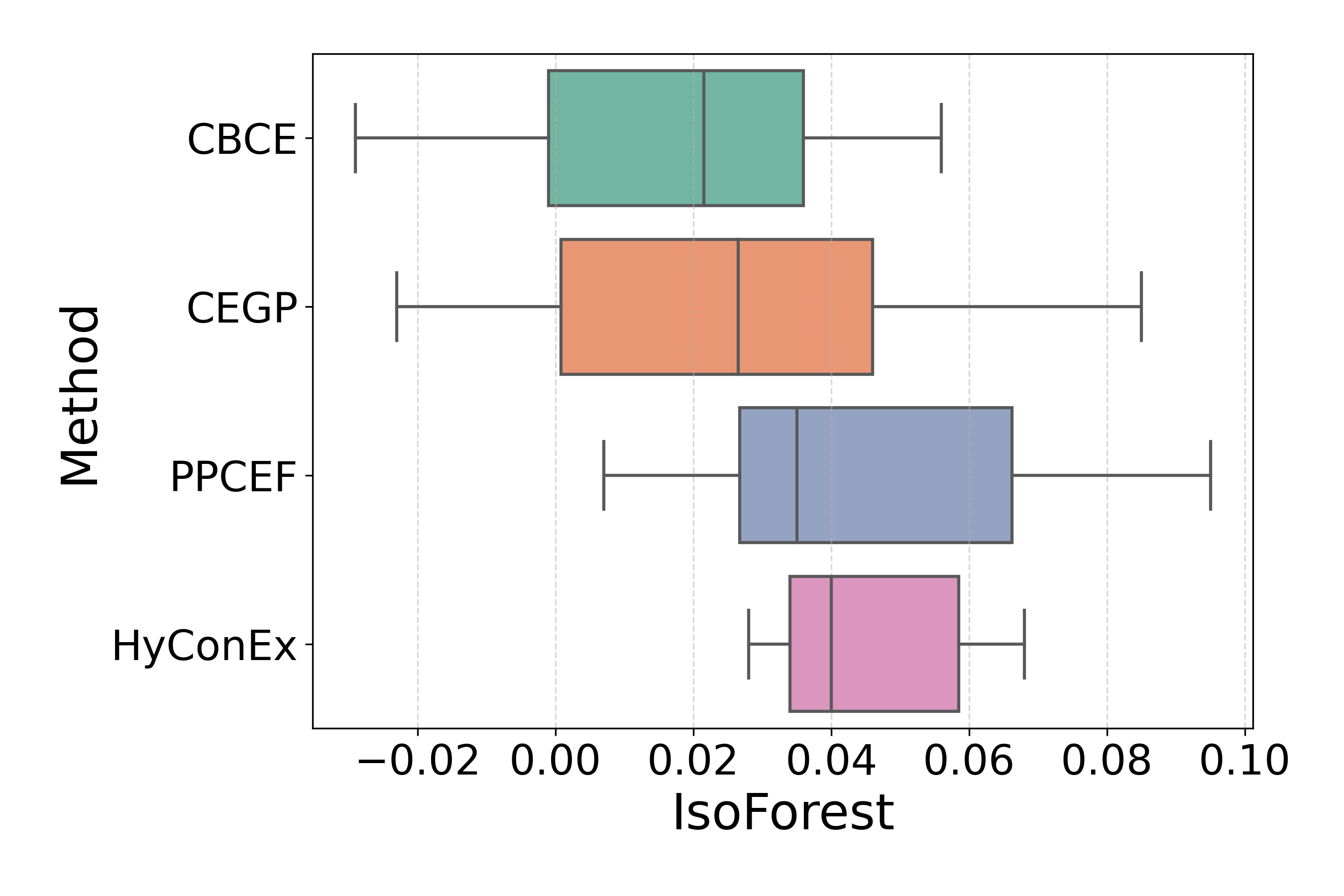} \includegraphics[width=0.32\linewidth, trim=35 0 30 0, clip]{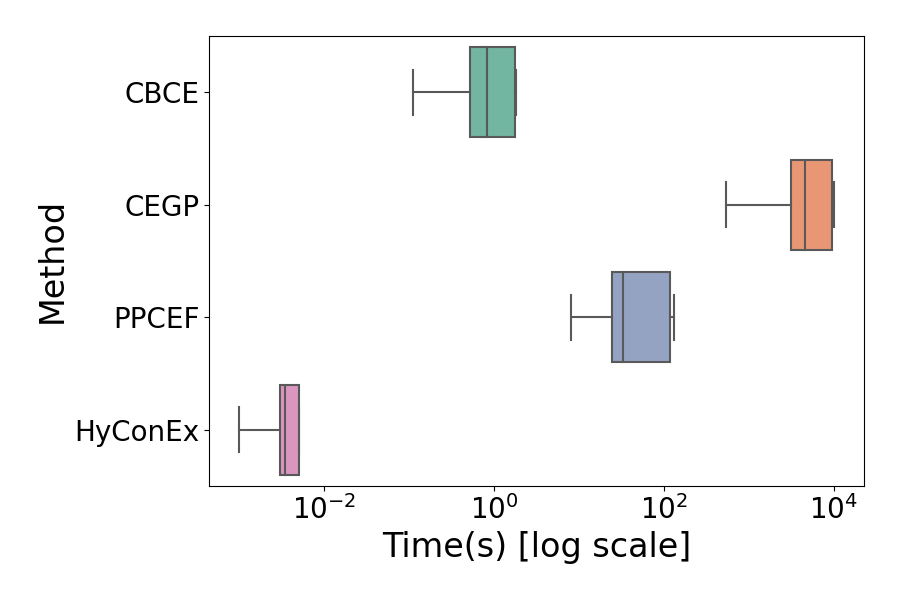}
    \caption{Summarized results from Tables \ref{table:countefacts_2_class}-\ref{table:countefacts_categorical} evaluating counterfactual explanations. We report only those methods which appear in all tables.}
    \label{fig:summary}
\end{figure*}

\paragraph{Results}

\begin{table*}[t]
\setlength{\tabcolsep}{2,75pt}
\centering
\caption{Comparative Results for Binary Classification Datasets}
\label{table:countefacts_2_class}
\centering
\footnotesize
\begin{tabular}{ll|ccccccccc}
\toprule
Dataset & Method & Cover.$\uparrow$ & Valid.$\uparrow$ & L1$\downarrow$ & L2 $\downarrow$ & P.Plaus.$\uparrow$ & LogDens$\uparrow$ & LOF$\downarrow$ & IsoForest$\uparrow$ & Time(s)$\downarrow$  \\
\midrule % 0.1
\multirow{6}{*}{Moons} & CBCE & 1.00 & \textbf{1.000} & 0.576 & 0.459 & \textbf{1.000} & \textbf{2.11} & \textbf{0.963} & \textbf{0.033} & 0.75 \\
& CEGP & 1.00 & \textbf{1.000} & \textbf{0.190} & \textbf{0.154} & 0.000 & -2.39 & 1.379 & -0.0006 & 4485.89 \\
& CEM & 0.99 & \textbf{1.000} & 0.543 & 0.508 & 0.059 & -6.740 & 1.786 & -0.077 & 2752.94 \\
& WACH & 0.99 & 0.975 & 0.208 & 0.178 & 0.000 & -1.921 & 1.300 & -0.001 & 2724.99 \\
& PPCEF & 1.00 & \textbf{1.000} & 0.337 & 0.261 & 0.990 & 1.32 & 1.032 & 0.022 & 35.62  \\
& \our{} & 1.00 & \textbf{1.000} & 0.502 & 0.409 & \textbf{1.000} & 1.93 & 0.985 & \textbf{0.033} & \textbf{0.003} \\
\midrule % 1.0
\multirow{6}{*}{Law} & CBCE & 1.00 & \textbf{1.000} & 0.895 & 0.597 & 0.493 & -0.45 & 1.219 & -0.024 & 1.73 \\
& CEGP & 1.00 & \textbf{1.000} & \textbf{0.204} & \textbf{0.170} & 0.457 & 0.95 & 1.042 & 0.049 & 10155.25 \\
& CEM & 0.99 & \textbf{1.000} & 0.310 & 0.297 & 0.518 & 1.19 & 1.088 & 0.005 & 5491.65 \\
& WACH & 1.00 & 0.995 & 0.412 & 0.341 & 0.511 & 1.16 & 1.084 & -0.002 & 5579.35 \\
& PPCEF & 1.00 & \textbf{1.000} & 0.338 & 0.202 & \textbf{1.000} & 1.83 & 1.032 & \textbf{0.067} & 131.00 \\
& \our{} & 1.00 & \textbf{1.000} & 0.359 & 0.227 & 0.894 & \textbf{2.54} & \textbf{1.026} & 0.060 & \textbf{0.005} \\
\midrule % 1.0
\multirow{6}{*}{Audit} & CBCE & 1.00 & \textbf{1.000} & 3.128 & 1.496 & \textbf{1.000} & 51.90 & \textbf{1.031} & \textbf{0.097} & 0.46 \\
& CEGP & 1.00 & \textbf{1.000} & 0.148 & \textbf{0.081} & 0.164 & 27.17 & 1.576 & 0.085 & 2578.10 \\
& CEM & 1.00 & \textbf{1.000} & \textbf{0.145} & 0.126 & 0.050 & -23.41 & 1.620 & 0.083 & 1502.50 \\
& WACH & 1.00 & \textbf{1.000} & 0.232 & 0.158 & 0.016 & -51.49 & 1.545 & 0.075 & 1586.24 \\
& PPCEF & 1.00 & \textbf{1.000} & 0.805 & 0.246 & \textbf{1.000} & 52.87 & 1.452 & 0.095 & 22.87 \\
& \our{} & 1.00 & 0.975 & 0.495 & 0.151 & 0.615 & \textbf{58.30} & 1.484 & 0.068 & \textbf{0.003} \\
\midrule % 1.0
\multirow{6}{*}{Heloc} & CBCE & 1.00 & \textbf{1.000} & 3.524 & 1.084 & \textbf{1.000} & 32.27 & \textbf{0.989} & 0.056 & 11.15 \\
& CEGP & 1.00 & \textbf{1.000} & \textbf{0.027} & \textbf{0.022} & 0.428 & 30.19 & 1.073 & 0.059 & 42475.70 \\
& CEM & 1.00 & \textbf{1.000} & 0.0447 & 0.043 & 0.361 & 28.86 & 1.079 & 0.053 & 25924.86 \\
& WACH & 1.00 & 0.998 & 0.345 & 0.193 & 0.031 & -49.78 & 1.189 & 0.025 & 26430.91 \\
& PPCEF & 1.00 & 0.993 & 0.457 & 0.124 & 0.999 & \textbf{34.93} & 1.042 & \textbf{0.074} & 486.02 \\
& \our{} & 1.00 & 0.997 & 0.416 & 0.142 & 0.602 & 34.61 & 1.073 & 0.065 & \textbf{0.019} \\
\bottomrule
\end{tabular}
\end{table*}

\begin{table*}[t]
\setlength{\tabcolsep}{2,75pt}

\centering
\caption{Comparative Results for Multi-Class Datasets (averaged over all classes)}
\label{table:countefacts_multi_class}
\centering
\footnotesize
\begin{tabular}{ll|ccccccccc}
\toprule
Dataset & Method & Cover.$\uparrow$ & Valid.$\uparrow$ & L1$\downarrow$ & L2$\downarrow$ & P.Plaus.$\uparrow$ & LogDens$\uparrow$ & LOF$\downarrow$ & IsoForest$\uparrow$ & Time(s)$\downarrow$  \\
\midrule % 1.0
\multirow{5}{*}{Blobs} & CBCE & \textbf{1.00} & \textbf{1.000} & 0.764 & 0.581 & 0.333 & 1.03 & 1.31 & -0.029 & 0.92 \\
& CEGP & 0.98 & \textbf{1.000} & \textbf{0.370} & \textbf{0.303} & 0.000 & -9.35 & 2.625 & -0.076 & 4607.08 \\
& WACH & 0.99 & 0.993 & 0.393 & 0.341 & 0.000 & -8.650 & 2.756 & -0.081 & 2380.44 \\
& PPCEF & \textbf{1.00} & \textbf{1.000} & 0.675 & 0.500 & \textbf{1.000} & 2.26 & 1.162 & 0.007 & 26.74 \\
& \our{} & \textbf{1.00} & \textbf{1.000} & 0.701 & 0.527 & 0.943 & \textbf{2.88} & \textbf{1.058} & \textbf{0.028} & \textbf{0.004} \\
\midrule % 1.0
\multirow{5}{*}{Digits} & CBCE & \textbf{1.00} & \textbf{1.000} & 15.448 & 3.002 & 0.330 & 53.65 & 1.073 & 0.029 & 1.80 \\
& CEGP & 0.96 & \textbf{1.000} & 6.090 & 1.423 & 0.011 & -392.35 & 1.201 & -0.023 & 8165.49 \\
& WACH & \textbf{1.00} & \textbf{1.000} & \textbf{2.076} & \textbf{0.999} & 0.018 & -69.30 & 1.097 & 0.005 & 4354.78 \\
& PPCEF & \textbf{1.00} & \textbf{1.000} & 6.550 & 1.130 & \textbf{1.000} & 55.98 & 1.055 & 0.029 & 86.34\\
& \our{} & \textbf{1.00} & 0.999 & 6.978 & 1.396 & 0.809 & \textbf{62.98} & \textbf{1.051} & \textbf{0.040} & \textbf{0.005} \\
\midrule % 0.1
\multirow{5}{*}{Wine} & CBCE & \textbf{1.00} & \textbf{1.000} & 3.560 & 1.178 & 0.667 & 7.25 & 1.062 & 0.037 & 0.11\\
& CEGP & \textbf{1.00} & \textbf{1.000} & 1.346 & \textbf{0.547} & 0.000 & -10.70 & 1.174 & 0.005 & 544.26 \\
& WACH & \textbf{1.00} & \textbf{1.000} & \textbf{1.010} & 0.625 & 0.015 & -12.379 & 1.228 & 0.005 & 287.38 \\
& PPCEF & \textbf{1.00} & \textbf{1.000} & 1.959 & 0.600 & \textbf{1.000} & 7.38 & 1.033 & \textbf{0.064} & 8.12  \\
& \our{} & \textbf{1.00} & \textbf{1.000} & 2.241 & 0.738 & 0.700 & \textbf{8.40} & \textbf{0.999} & 0.054 & \textbf{0.001}  \\
\bottomrule
\end{tabular}
\end{table*}

\begin{table*}[t]
\setlength{\tabcolsep}{2,5pt}

\centering
\caption{Comparative Results for Datasets with Categorical Features}
\label{table:countefacts_categorical}
\centering
\footnotesize
\begin{tabular}{cl|ccccccccc}
\toprule
Dataset & Method & Cover.$\uparrow$ & Valid.$\uparrow$ & L2$\downarrow$ & Ham.$\downarrow$ & P.Plaus.$\uparrow$ & LogDens$\uparrow$ & LOF$\downarrow$ & IsoForest$\uparrow$ & Time(s)$\downarrow$  \\
\midrule
\multirow{4}{*}{\rotatebox{90}{Adult}} & CBCE & \textbf{1.00} & \textbf{1.000} & 0.334 & 0.652 & 0.751 & \textbf{30.79} & \textbf{0.999} & -0.002 & 88.53 \\
& CEGP & 0.43 & 0.999 & 0.313 & \textbf{0.000} & 0.432 & 25.05 & 1.777 & \textbf{0.037} & 67348.95 \\
& PPCEF & \textbf{1.00} & 0.672 & 0.379 & 0.249 & 0.646 & 27.14 & 1.599 & 0.031 & 1453.48 \\
& \our{} & \textbf{1.00} & \textbf{1.000} & \textbf{0.136} & 0.328 & \textbf{0.861} & 29.86 & 1.257 & \textbf{0.037} & \textbf{0.05} \\
\midrule
\multirow{4}{*}{\rotatebox{90}{Credit-G}} & CBCE & \textbf{1.00} & \textbf{1.000} & 0.944 & 0.584 & \textbf{0.517} & \textbf{18.83} & \textbf{1.007} & 0.014 & 0.51 \\
& CEGP & \textbf{1.00} & \textbf{1.000} & \textbf{0.311} & 0.026 & 0.258 & 0.45 & 1.029 & 0.024 & 3035.62 \\
& PPCEF & \textbf{1.00} & 0.517 & 0.387 & \textbf{0.005} & 0.358 & 4.69 & 1.023 & 0.026 & 23.45 \\
& \our{} & \textbf{1.00} & \textbf{1.000} & 0.585 & 0.067 & 0.492 & 8.51 & 1.022 & \textbf{0.031} & \textbf{0.003} \\
\midrule
\multirow{4}{*}{\rotatebox{90}{Credit-A}} & CBCE & \textbf{1.00} & \textbf{1.000} & 7.675 & 0.486 & 0.478 & 10.478 & 1.208 & 0.002 & 0.58 \\
& CEGP & 0.96 & \textbf{1.000} & \textbf{0.372} & \textbf{0.000} & 0.106 & 7.888 & 1.048 & 0.029 & 3415.51 \\
& PPCEF & \textbf{1.00} & 0.355 & 0.669 & 0.082 & 0.167 & 13.853 & \textbf{1.040} & 0.039 & 29.94 \\
& \our{} & \textbf{1.00} & \textbf{1.000} & 1.606 & 0.164 & \textbf{0.638} & \textbf{21.749} & 1.055 & \textbf{0.040} & \textbf{0.003} \\
\bottomrule
\end{tabular}
\end{table*}

Our experiments demonstrate that \our{} consistently achieves near-perfect coverage and validity across diverse settings while offering substantial computational advantages (see Tables \ref{table:countefacts_2_class}-\ref{table:countefacts_categorical}). Results presented in Table \ref{table:countefacts_2_class} show that for binary classification tasks, our method generates counterfactual explanations with competitive proximity metrics and high plausibility, processing the Moons dataset in merely 0.003 seconds compared to 2724.99 seconds for Wachter. CEGP offers excellent L1 and L2 proximity, though this comes at the expense of lower probabilistic plausibility and density. Conversely, CBCE is very fast and generally achieves good LOF and IsoForest outlier scores, but it consistently yields the worst outcomes on distance-based metrics. All methods among CEGP, CEM, and Wachter require hours of optimization to generate counterfactual examples for all points in the test set, making them impractical for real-time applications. PPCEF specializes in maximizing P.Plaus., and its computational cost remains relatively moderate.

In the multi-class case (\(K\)-class scenario), we consider a setting where, for a given factual point, we create \(K-1\) counterfactual points corresponding to all available target classes. Table~\ref{table:countefacts_multi_class} presents the resulting metrics aggregated over all classes (we omit the CEM method, as it does not support targeted counterfactuals).
Noteworthy findings in the multi-class case, beyond those reported in Table~\ref{table:countefacts_2_class}, include the competitive performance of the Wachter method on L1 and L2 metrics, as well as the top scores achieved by \our{} on LogDens, LOF, and IsoForest. Our main advantage—time efficiency—becomes even more important here, as the number of possible alternative classes increases (e.g., up to 9 in the case of \textit{Digits}).

The comparison on datasets with categorical features is illustrated in Table~\ref{table:countefacts_categorical} (we exclude the Wachter model since it is not designed for datasets with categorical features). Although both \our{} and PPCEF use INFs to estimate data density, only our method does it correctly for categorical variables (compare validity scores and see Section~\ref{sec:cat} for details). \our{} once again maintains strong performance, generating valid counterfactuals that require only minimal modifications and remain within high-density regions of the training data distribution. Among the remaining approaches, an interesting observation can be made for CEGP, which aggressively minimizes the Hamming distance and deteriorates other metrics (see low coverage for the Adult dataset).

The results aggregated over all datasets from Tables~\ref{table:countefacts_2_class}-\ref{table:countefacts_categorical} are shown in Figure~\ref{fig:summary}. Notably, \our{} achieves very good results on in-distribution metrics (P.Plaus., LogDens, LOF, IsoForest) being significantly faster than other methods. Nevertheless, the proximity of counterfactuals generated by \our{} is frequently lower than the ones produced by competitive methods. 

\subsection{Ablation Study}

\begin{table*}[t]
\setlength{\tabcolsep}{2pt}

\centering
\caption{Ablation Study: Importance of Loss Components on the Heloc Dataset. Best results are in \textbf{bold} and the second best are \underline{underlined}.}
\footnotesize
\label{table:ablation_importance}
\begin{tabular}{l|cccccccc}
\toprule
Loss Type & Cover.$\uparrow$ & Valid.$\uparrow$ & L1$\downarrow$ & L2$\downarrow$ & P.Plaus.$\uparrow$ & LogDens$\uparrow$ & LOF$\downarrow$ & IsoForest$\uparrow$ \\
\midrule
Base                & \textbf{1.00} & 0.000        & -           & -           & -           & -             & -           & -          \\
Base+CE             & \textbf{1.00} & \textbf{1.000}  & 6.394       & 1.628       & 0.000       & -34761532.00 & 3.925       & -0.036     \\
Base+CE+Flow        & \textbf{1.00} & \textbf{1.000}  & 1.461       & 0.475       & \textbf{0.662} & \underline{33.707}  & 1.108       & 0.058      \\
Base+CE+Dist        & \textbf{1.00} & 0.988        & \textbf{0.098}  & \textbf{0.032}  & 0.394       & 27.66       & \textbf{1.072}  & \underline{0.060} \\
Full                & \textbf{1.00} & \underline{0.997}  & \underline{0.416}  & \underline{0.142}  & \underline{0.602}  & \textbf{34.61}  & \underline{1.073}  & \textbf{0.065} \\
\bottomrule
\end{tabular}
\end{table*}

To evaluate the contribution of each loss component, we performed an ablation study on the \textit{Heloc} dataset. Table~\ref{table:ablation_importance} reports the performance metrics for the following loss configurations:
\begin{compactitem}
    \item \textbf{Base:} The loss contains only the cross-entropy of the input data, which fails to produce valid counterfactuals.
    \item \textbf{Base+CE:} Augments the base loss with a cross-entropy (CE) term for counterfactual example. This setup achieves perfect validity but yields counterfactuals with poor plausibility and highly negative log density values, suggesting out-of-distribution samples.
    \item \textbf{Base+CE+Flow:} Adds a flow constraint to the Base+CE loss to encourage plausibility. It retains perfect validity while significantly boosting plausibility.
    \item \textbf{Base+CE+Dist:} Integrates a distance minimization (Dist) term with the Base+CE loss to enforce proximity, leading to a dramatic reduction in L1 and L2 distances at the cost of a slight decrease in validity.
    \item \textbf{Full:} Combines all components (CE, Flow, and Dist) to balance validity, proximity, and plausibility.
\end{compactitem}
The results demonstrate that each component contributes to the quality of generated counterfactuals, with the full model effectively balancing the trade-offs.

\section{Conclusion}
%%%%%%%%%%%%%%%%%%%%%%%%%%%%%%%%%%%%

Understanding model decisions is crucial in explainable AI, mainly through counterfactual explanations highlighting key feature modifications affecting predictions. While existing methods generate counterfactuals for external models, no interpretable classifier has produced predictions and counterfactuals simultaneously. We introduced \our{}, a deep hypernetwork-based model tailored for tabular data, which provides reasonable  classifications and generates counterfactual examples within high-density class regions. Our approach achieves competitive performance while ensuring interpretability, making \our{} a unique all-in-one neural network that seamlessly integrates prediction and explanation.

\paragraph{Limitations} \our{} is intentionally tailored for tabular datasets. This specialization yields strong performance where interpretability is most critical, but broadening the method to other data modalities would require additional architectural modifications. Although inference is highly efficient, this capability depends on an upfront training phase; once completed, however, \our{} supports real-time deployment and substantially outperforms prior methods in runtime.

\our{} delivers a single counterfactual candidate for every alternative class. While this approach allows us to return a plausible counterfactual example with good proximity, we do not provide a user with multiple candidates as in the case of generative models. Moreover, since counterfactual explanations are embedded into \our{} classifier, we do not take into account the change of the classification decision boundary. 
This property refers to the ability of counterfactual examples to withstand small perturbations in the classifier. 
Additional modifications of \our{} would be required to produce counterfactuals, which are robust to such modifications. 

\paragraph{Future work}

The concept of a hypernetwork generating local linear decision boundaries, introduced by IMN and further developed in \our{}, demonstrates the considerable potential of this type of architecture. Therefore, future work on this framework can progress through both incremental refinements and more transformative innovations.

In particular, it is worth to consider different hypernetwork backbones, such as transformer-based (e.g., FT-Transformer~\cite{gorishniy2021revisiting}), retrieval-augmented (TabR~\cite{gorishniy2023tabr}) or ensemble-like (TabM~\cite{gorishniy2024tabm}) models.  Additionally, we could potentially try using other differentiable density estimation models, especially for high-dimensional or categorical datasets. Finally, it would be valuable to extended the scope of \our{} to other modalities, such as images or texts.

\section*{Acknowledgment}

The research of P. Marszałek, U. Movsum-zada supported by the National Science Centre (Poland), grant no. 2023/50/E/ST6/00169. The work of K. Książek was supported by the flagship project ''Artificial Intelligence Computing Center Core Facility'' from the DigiWorld Priority Research Area within the Excellence Initiative – Research University program at Jagiellonian University in Krakow. The work of P. Spurek was supported by the National Science Centre (Poland), Grant No. 2023/50/E/ST6/00068.  The work of M. \'Smieja was supported by the National Science Centre (Poland), Grant No. 2022/45/B/ST6/01117. Oleksii Furman work was supported by the National Science Centre (Poland) Grant No. 2024/55/B/ST6/02100.

%Bibliography

\end{document}